# Assessment of cognitive characteristics in intelligent systems and predictive ability

Oleg V. Kubryak[1,2], Sergey V. Kovalchuk[3], Nadezhda G. Bagdasaryan [4]

[1] National Research University "Moscow Power Engineering Institute", Moscow, Russia
[2] Anokhin Institute of Normal Physiology, Moscow, Russia
[3] ITMO University, Saint Petersburg, Russia
[4] Bauman Moscow State Technical University, Moscow, Russia

**Abstract**. The article proposes a universal dual-axis intelligent systems assessment scale. The scale considers the properties of intelligent systems within the environmental context, which develops over time. In contrast to the frequent consideration of the "mind" of artificial intelligent systems on a scale from "weak" to "strong", we highlight the modulating influences of anticipatory ability on their "brute force". In addition, the complexity, the "weight" of the cognitive task and the ability to critically assess it beforehand determine the actual set of cognitive tools, the use of which provides the best result in these conditions. In fact, the presence of "common sense" options is what connects the ability to solve a problem with the correct use of such an ability itself. The degree of "correctness" and "adequacy" is determined by the combination of a suitable solution with the temporal characteristics of the event, phenomenon, object or subject under study.

## Introduction

Let's suppose that the definitions of "artificial intelligence", "power of intelligence" and "critical thinking" are unambiguous, that these terms are agreed upon and accepted by everyone, and are understood by everyone to the same extent as, for example, the definition of "light". Our further reasoning is based on the general idea that the manifestations of an entity that occupies thoughts and is the subject of people's real activity can characterize this entity in sufficient detail to attempt to isolate, study and describe it. In conversations with people of past generations, a frequent response to a question would be: "I don't know enough about this", suggesting to ask someone else who, in the opinion of the respondent, might know better and give the correct answer. Such a critical approach to assessing one's own "intellectual power" is traditional and is reflected in the most ancient, millennia-old evidence [1] : different peoples' fairy tales. For example, in the Russian folk tale "Rejuvenating apples and life-giving water", the main character seeks advice in solving a difficult matter from a magical assistant, Baba Yaga: "give your head to my mighty shoulders, direct me to mind and to reason". Today one may get the impression that some "magic assistant", for example, a search engine on the Internet, is always nearby and thereby guarantees omniscience, or at least a self-evident opportunity for anyone to have a genuine opinion and freely judge this or that event or phenomenon, complex system or scientific work, even without resorting to the "magic assistant". In other words, the availability of knowledge, as it were, makes the person next to you a priori "smarter". It is likely that when studying this kind of phenomenon, we are dealing not just with a lack of time for reflection due to the high pace of life, and not with a manifestation of insufficient attentiveness, since attempts to increase "mindfulness" do not enhance critical thinking [2]. Also, the presence of a large amount of knowledge does not make a person smarter - let's remember another old Indian tale about "foolish smart men", the three Brahmins who "studied many different sciences and considered themselves smarter than everyone" (quote from the Russian translation[3]). Having invited a "stupid" peasant to guide them through the mountains, they found a dead lion and revived him, despite the warnings of the guide, who was the only one to escape from the resurrected predator. The tale ends with this message: "To a fool, science is like the light of a lamp to a blind man!".

Describing his theory of functional systems, P. Anokhin noted that "any fractional function of an organism turns out to be possible only if, at the moment of the formation of a decision and a command to action, a prediction mechanism is immediately formed. It is quite obvious that machines that could ‘look into the future’ at every stage of their operation would receive a significant advantage over the modern ones" [4]. The idea of "prediction" as a distinctive feature for the creation of intelligent systems, also expressed by Norbert Wiener for cybernetics , was actively developed in the middle of the 20th century [6]. It preceded many modern problems of artificial intelligence construction, including the search for analogies with the living brain [7]. Once we focus on the ability to "see the future", it becomes clear that the "foolish smart men" from the Indian fairy tale were deprived of it, unlike the "stupid" peasant. In other words, neither knowledge, nor "big data" by itself determines critical thinking for a person or a similar property for artificial intelligence.

Thus, considering only computing power and performance as indicators of power of intellect is futile if predictive ability is ignored. An obligatory condition for the existence of real (living) intelligent systems is the consideration of time, its flow and the correlation of past, present and future events with the system itself and the surrounding world; a simplified diagram is presented in Figure 1. In the figure, an intelligent agent has access to sufficiently powerful intellectual tools that significantly enhance their ability to assess the environment (physical, social, informational), as well as their future states and associated risks. The agent also has access to tools for meta-cognitive analysis and assessment of the agent, their knowledge and abilities. However, access to these tools is limited by the cognitive interaction capabilities of an intelligent agent.

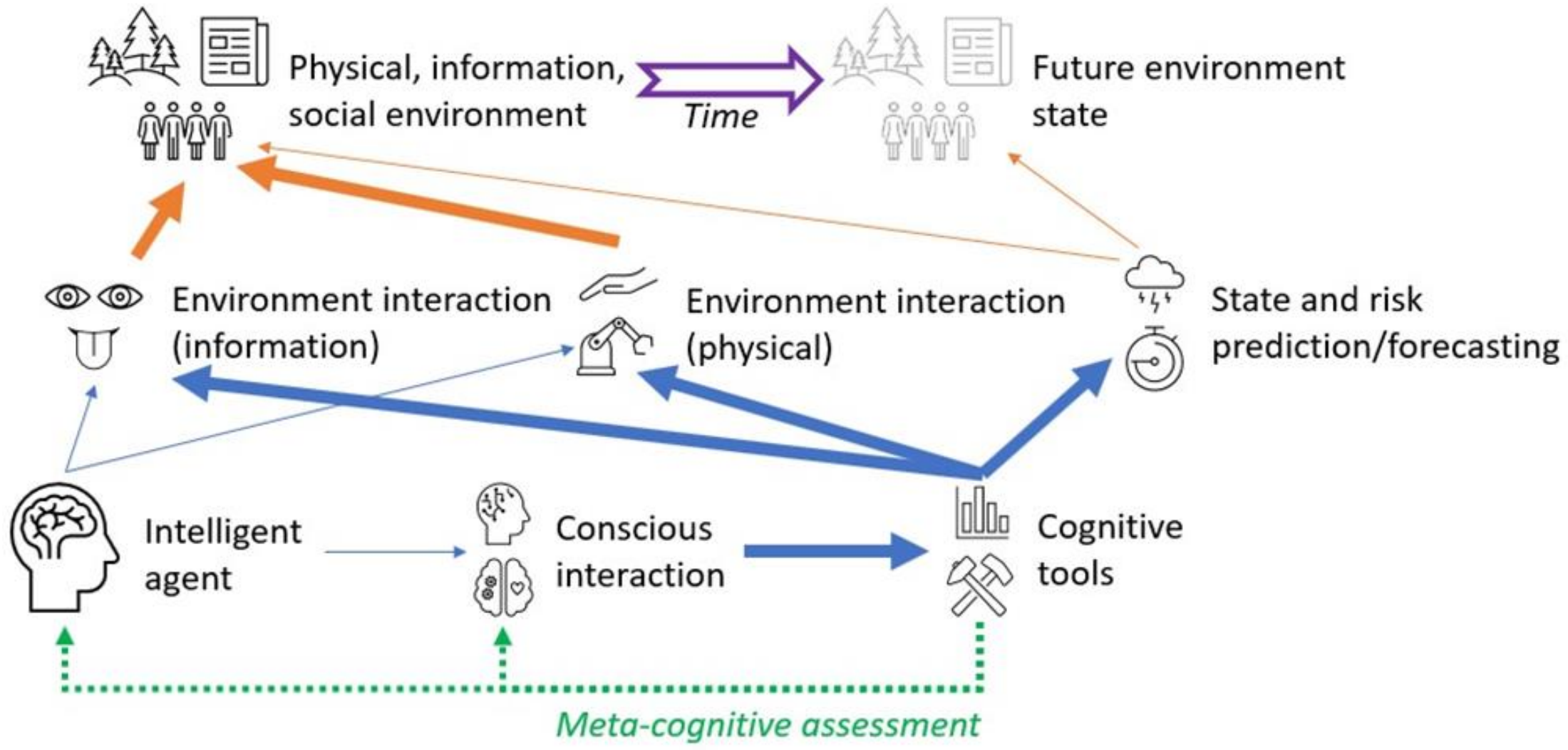

Figure 1 – The interaction of various conditions taken into account and used by the intelligent system

In this context, we believe that the assessment of the power of intellect should not be a linear scale from strong to weak, but should rather contain at least two axes: the scales of brute force and smart force along with the scale of adequacy. Therefore, the purpose of this article is to present our approach to the assessment of intelligent systems.

## The weight of objects and the complexity of intellectual tasks

If you ask an ordinary person whether they can lift a weight of 5 kilograms, they will most likely answer affirmatively, and truthfully.. And honest. If you repeat the question regarding progressively greater weights, reaching, for example, up to 100 or 200 kilograms, you will see that people usually judge the proportionality of their own physical strength very well. There is a clear understanding that a certain weight cannot be lifted. It is much more complicated when one is asked to evaluate events or phenomena. If you offer someone who does not know mathematics to solve several equations, or someone who does not know a programming language to write code, then the excessive complexity of the task will be obvious. However, reasoning about extremely complex things that do not provide such explicit formal markers of "complexity" (as program code or equations) can be assessed by a person as accessible to their intellectual "power". The topics on which many people easily express judgments include, for example, ideas about the work of the government, about epidemiological measures and vaccination [8,9], etc. Thus, without the "magic veil" of formulas, equations and codes, complex cognitive tasks often seem quite clear and feasible for analysis, for having an opinion or a judgment.

In such situations, the fallacy of judgments about complex things and the low degree of critical thinking can be compared (in an exaggerated way) to what happens in mental disorders: "delusion does not necessarily have to come from an external event or be centered around it. On the contrary, delusions can start as a thought or other internal event, such as a hallucination, develop into a deeply held opinion, suggestion, or judgment about oneself or one's status. In sum, delusion begins with a hasty judgment about an object of personal significance, which is then developed and brought to the point of absurdity by further hasty judgments" [10]. It should also be noted that the theme of madness and rationality is also reflected in philosophy, for example, in the works of Michel Foucault, contrary to a more traditional view of psychiatry and questions of the mental norm (normal intellectual process) [11]. Thus, the most appropriate explanation of the problems of critical thinking and evaluation of events today is correlating the probability of events (facts) with existing experience (previously obtained representations, opinions or delusions), while ranking the degrees of confidence. We believe that it is possible to describe this formally, for example, using Bayesian approaches [12]. Another approach is through catastrophe theory [13], in which the "breakdown" of critical thinking can be likened to a sudden (for an external observer) breakage of the ruler on which the load is placed, or the theory of self-organized criticality [14], in which the "breakdown" is due to the previous "accumulation" of small intellectual errors. This is reflected in the characterization of any unreliable statement as "built on sand".

## Nonlinearity and elements of artificial intelligence

In a chaotic system, the initial condition always leads to the same final state, following a certain law, while in a stochastic system the initial condition can lead to a variety of different final states [15]; it is also challenging to differentiate between random and non-random events [16]. Therefore, there can be different approaches to the design of artificial intelligence. Today, AI is undergoing a phase of active development, often associated with deep learning technologies. The success of deep learning in a number of areas (e.g, computer vision, speech recognition) promotes the use of these methods for various applications. Nevertheless, AI models, like any other models used in an applied setting, exhibit a high degree of uncertainty in different aspects [17]. Moreover, many AI models (especially those formed within the connectivist framework) display the properties of a "black box". An increase in the size of models in learning problems leads not only to greater computational complexity, but also to the unstable operation of learning algorithms (the so-called "curse of dimensionality"), which further increases the uncertainty of the structure and parameters of models. As a result, the generated models require the development of mechanisms for interpreting and explaining both the results of predictive modeling and the structure of the generated model. In the current AI paradigm, such mechanisms appear in the areas of explicable AI (eXplainable AI, XAI) and interpretable machine learning [18,19], which focus on structuring models and simulation results, while taking into account the interpretable contribution of individual (subject-interpreted) factors. Mechanisms of Bayesian inference [20,21], and causal methods developed by J. Pearl [22] and other authors [23] also explicitly reveal the structure of models. Such approaches allow a shift to a new, "metacognitive" level of working with AI: using AI methods to identify the structuring of its own characteristics and reducing internal structural and parametric uncertainty.

On the other hand, the technical closedness of the model ("black box") may bring limitations related to understanding and willingness to use the results of AI work on the part of end users. These issues include issues of trust [24], morality [25] and ethics [26] of AI. The problem of explainability is expanded by questions of unbiasedness [27], stability [28], conformity to values and norms (value-alignment) [29], relationship with the objective formulation of the problem being solved [30] etc. There is a transition to the extended tasks of AI development: instead of creating isolated solutions that replace humans, the task is to organize a collaboration between humans and AI within different approaches, such as human-centered AI (human-centered AI) [31,32], and augmented intelligence [33]. It is also critical to build effective interactions and mutual understanding between humans and AI, which leads to a rethinking of both technical and logical means of information exchange. For example, there are new perspectives on the issues of visualization [34] and interaction in natural language [35], issues of automating the management of AI models [36] and the relationship of understanding user experience (user experience, UX) in light of explainable AI technologies [37]. Thus, the "metacognitive" level is already attained through the human-AI system, in which the development of an isolated autonomous (albeit detailed) AI system is not the goal, but only a tool for achieving effective synergy of system elements.

Hence, the complex combination of different elements into an intellectual system further complicates it, increasing non-linearity, or, at least, moving further away from the "simple", "formal" sequence of development of the cognitive process, which, as it may seem to the observer, corresponds to "ordinary", "understandable" logic. In other words, the method of solving a problem, reminiscent of "insight" or "intuition", marks the high nonlinearity of the intellectual system, which is associated with effective intellectual work. Metaphorically such a property of an intellectual system can be called "hyperdrive" or a "hypercognitive" property, like the engines of interstellar ships described in fantasy works, and the method of solving a problem is similar to traveling through a "wormhole" in space.

## Cognitive limitations and their connection to ethics in artificial intelligence

Kurt Vonnegut wrote: "If it weren't for [people],the world would be an engineer's paradise." It is precisely people that should be in focus when it comes to artificial intelligence. All procedures, from data entry to interpretation of results, require the algorithmization of critical thinking, which takes cognitive factors into account. These factors include dominant paradigms, concepts, traditions, beliefs, social myths and cultural stereotypes, and all of them must be considered in creating strong AI. Recall that the features of cognitive systems can be considered both through the humanistic approach (C. Rogers' law of congruence), and in light the cognitive approach. Furthermore, other aspects of human intelligence should be taken into account, for example, consciousness (being receptive to the environment); self-awareness (being aware of oneself as a separate person, in particular, to understand one's own thoughts; empathy (the ability to "feel", emotional intelligence); wisdom, motivation, etc. The complexity of social structure is accompanied by a high degree of stochasticity of global processes, the high level of risk in any segment of society. The conditions of uncertainty and weakly expressed determinism form a space of non-standard problems that require appropriate, equally non-standard approaches and solutions within the framework of strategic forecasting. One should not ignore the possibility of singularity, which is very characteristic of social processes, wherein one single event generates a new meaning [38]. This is similar to mathematics, where singularity is understood as a point at which a function approaches infinity or becomes unpredictable.We assume that at the moment of the onset of singularity, usual laws cease to operate.[39]

Developing forecasts at bifurcation points or in periods of catastrophe, when systems that have exhausted their evolution, fall apart, is especially difficult. Can artificial intelligence, trained to detect the "correct" patterns, and acting according to the input data, offer an adequate exit vector from bifurcation points? Furthermore, the current state of mankind, described by A.P. Nazaretyan as "polyfurcation", surpasses previous phase transitions in the history of the Earth in planetary and, possibly, cosmic consequences [40]. The result of polyfurcation is the formation of "hybrid structures". The concept of hybridity has become popular today when explaining complex systems of different levels. The term itself has acquired a universal meaning and is used in a variety of contexts: from political "hybrid" regimes to "hybrid" forms of education, to mixed reality technologies combining the real and virtual worlds, and to cognitive hybrid intelligent decision support and forecasting systems. The latter refers to a hybrid of the fuzzy cognitive map (FCM; this map reflects the analysis of the consciousness of people involved in the analyzed processes) and the neuro-fuzzy network ANFIS, which supports decision-making in dynamic unstructured situations, for example, the irrational behavior of people, or, in the present work, in behavioral economics. [41].

For us, the research can be a detailed case study, because it provides an example of taking into account the influence of social, cognitive and emotional factors on economic decision making. It also becomes clear that approaches from neurophysiology, psychology, linguistics, anthropology, as well as the entire apparatus of modern computer science, robotics and brain modeling are in demand when solving this type of problem. All these areas are connected with the representation of knowledge in the human brain. The decision support in unstructured situations model is interesting, because it makes it possible to evaluate, on the basis of a scale of criteria, the weight of a number of alternatives in forecasts for the development of a nonlinear situation, under conditions of uncertainty, stochasticity, limited information, or weak determinism. It is in these cases that fuzzy logic (introduced in 1965 by Professor Lotfi A. Zadeh at the University of California at Berkeley) is reasonably used as a structural element of post-non-classical mathematics. Linguistically, it can be expressed in "yes, no, maybe", all in a single utterance during reasoning. The advantages of the method are flexibility, the ability to set the rules and accept even inaccurate, distorted and erroneous information; all of this is inherent in human reasoning.

In other words, the choice of a situational strategy and strategic forecasting today are unthinkable without the widespread use of neural programming, which has been studied in detail, while the choice of the most effective strategy is impossible without the use of fuzzy logic methods.

Let us return, however, from long-term global forecasting to solving utilitarian, pragmatic tasks. These tasks nevertheless require the development of certain strategies, which increasingly resort to AI. Until recently, it was believed that it was impossible to entrust AI withrecruiting, training and retaining talents. First, this procedure requires great responsibility: the selection of personnel is the key to business success. Any great business strategy is just as bad as not having one if the wrong person is put in the wrong place. However, it is not so much the responsibility of those who deal with human resources, but the complexity of the task, which determines the competence of applicants based on dimensional values, such as the level of knowledge, possession of professional skills, self-motivation, self-esteem and perception of roles. It is required to carry out an evidence-based assessment of weakly formalized, or even non-formalized cognitive tasks in order to predict the qualities of the applicant, while one can only count on the evidence of the correctness of the solution in well-formalized tasks (for example, mathematical ones). Meanwhile, in the literature there is a description of the development experience based on a quantitative mathematical model of a predictive model that uses AI to determine the level of personnel competence to optimize their performance. The model tests discrete and continuous relationships between competency dimensions and their synergistic effects. Four AI algorithms are used to train a dataset containing 362 data items consisting of two optimized combinations of dimensional values in step 1 and 360 data items derived from a dimensional relationship that produces synergistic effects. The proposed model predicts the competence of applicants and compares this competence with the value, while providing an optimized labor productivity for an adequate selection of human resources [42].

Although academic research in the field of intelligent automation is growing noticeably, there is still no understanding of the role of using these technologies in human resource management (HRM) at the organizational (firms) and individual (employees) level. In this regard, it is interesting to systemize academic contributions to intelligent automation and identify its problems for HRM. The study provides a review of 45 articles examining artificial intelligence, robotics, and other advanced technologies in an HRM environment, selected from publications in leading HRM, International Business (IB), General Management (GM), and Information Management (IM) journals. The results show that intelligent automation technologies represent a new approach to managing employees and improving firm performance, accompanied by an analysis of significant problems not only at the technological, but also at the ethical levels. The impact of these technologies is focused on human resource management strategies: changing the job structure, the collaboration between humans, robots and AI, the ability to make adequate decisions in recruitment, as well as training and evaluation of their production activities.[43]

Many problems arise in the use of biometric technologies that develop digital representations of bodily characteristics for identification of persons. Their widespread use has led to the institutionalization of registration and identification of persons, primarily in the field of the right to cross borders. Given the ever-increasing intensity of human movement (hopefully, only temporarily limited by the pandemic) around the globe for various purposes (tourism, various kinds of migration, academic exchanges, etc.), biometric technologies will soon take a dominant place in official registration procedures. However, as usual, technologies outrun the timely comprehension of their place in society, and it is this place that determines how digital procedures provide an objective and undeniable identification of a person. One of the first socio-anthropological studies based on ethnographic field work among technical developers, border police, forensic scientists, IT hacktivists and migrants emphasizes that in practice biometric technologies (as well as any other technologies) are embedded in specific social contexts, fraught with ambiguity and uncertainty, and are highly dependent on human interpretation and social identification.[44]

In his lecture at the St. Petersburg International Economic Forum, the world-famous popularizer of science, futurist and physicist Michio Kaku emphasized that "robots and artificial intelligence have three limitations. The first is interaction with people. Any work related to communication and trusting relationships requires human participation. The second is the recognition of different models and templates. Robots are bad at this. For example, if there is a table in front of the robot or it sees an image of a table, it will see some squares or triangles, but will not understand that this is a table, because it does not understand at all what a table is, what is the point of this object. Therefore, the robot needs a special program that will explain to him what a table is. And finally, the third is common sense. The robot does not have it. For example, you know that water is wet, not dry. Or that you can touch strings and they will react to your movements. Or that mothers are older than daughters. How do you know all this? This is the common sense that you have, robots don't. Even a five-year-old child understands these issues better than a robot" [45].

In addition to the objective limitations of AI capabilities, questions of ethical, moral and philosophical limitations are being raised more and more. Decisions made by artificial intelligence (no matter what criteria of social good they are guided by) will have unobvious consequences related to the specifics of its implementation and people's reactions to new technologies [46].

## Intelligence and predictive ability

The modern view of the "anticipatory" activity of the living brain as a manifestation of mental activity, [47], even in simple living beings (previously also proposed by L.V. Krushinsky [48], highlights the importance of this component of intelligence. The other side of the issue is an attempt at a cybernetic approach to living things, a view of "anticipation" as an element of self-regulation of complex systems. For example, the use of feedback and memory for medicine [49]. In Soviet systemic biological theories, such as, for example, the "theory of functional systems" by P.K. Anokhin [50] or the concept of motion control by N.A. Bernstein [51], the idea of anticipation, the "prediction of reality" is also key to explaining behavior.

Even "simple" living creatures, not usually considered particularly "intelligent", for example, fish, exhibit a wide range of complex behavior [52]. In other words, not only the total "strength of intellect" of higher animals plays an important role in behavior. There are other important features. In our opinion, one of them is the ability to predict, closely related to the ability to adapt to environmental conditions. The fact that such everyday adaptation is successfully carried out not only by conditionally "simple" living creatures like fish, but even by jellyfish and other, even simpler living creatures that do not have a brain similar to humans, indicates that it is not a single "computing power" that determines efficient behavior.

The cognitive abilities of both living organisms and AI systems differ in the complexity of the functional characteristics available to a single living or artificial agent. In systemizing of cognitive abilities, there are a number of solutions that formalize the relation of ordering (cognitive "superiority") depending on the set of available cognitive functions: these are cognitive scales.

Some examples of such scales include the BICA*AI community scale of cognition levels [53], выделяющая пять уровней от рефлексивного (reflexive) до метакогнитивного самосознания (meta-cognitive and self-aware), which distinguishes five levels from reflexive to metacognitive self-awareness (met-cognitive and self-aware), as well as the ConScale scale [54], based on the systematization of living organisms in terms of the level of cognitive development: from the uncontrolled behavior of an individual agent to "superconsciousness" (an agent operating with several "streams of consciousness"). In most cases, the highest levels of consciousness are characterized not only by predictive and proactive abilities, but also by the ability for (self-)reflection and meta-cognitive reasoning (i.e., conscious manipulation of one's own cognitive functional characteristics).

At the same time, to understand the structure of cognitive characteristics that explicitly configure the functional capabilities of consciousness, it is necessary to systematically structure their contents, features, interaction, etc. During the last decade of the development of cognitive science and AI, the curiosity towards this vector of study created many cognitive architectures [55] which formalize consciousness from different points of view, primarily, from the point of view of building AI. In general, approaches to the construction of cognitive architectures are organized within two directions: symbolism, which deductively structures the key elements of consciousness, and the emergent approach, most often within the framework of connectivism, which inductively generates elements of consciousness based on the combination of atomic elements.

One of the main current problems in the field of cognitive architectures is working with functional elements of consciousness that remain difficult to systematize even at the level of metacognitive consciousness, for example, intuition, solving creative problems, inspiration, etc. A number of developers of cognitive architectures and individual AI solutions offer original approaches for the functional reproduction of such aspects of human consciousness: the Clarion cognitive architecture defines creative work as the interaction of explicit and implicit knowledge [56], the Darwinian Neurodynamics architecture proposes the use of evolutionary approaches to generate new ideas [57], the developers of the well-known AlphaGo solution for playing Go, see reinforcement learning as a way to formalize intuition and the "sense of Go". However, these questions generally remain open despite the fact that many people actively use their own creative abilities. As a result, from the point of view of structuring the cognitive abilities of a natural intelligent agent (and, after that, an AI agent), there are at least two aspects that require systematization: the tools available to the agent (memory, thinking, will, ability to predict, make decisions etc.) and the "accessibility" of these tools to consciousness.

In this situation, metacognition becomes not just a tool for self-assessment, but a measurement that determines the ability of an agent to consciously and purposefully work with its functional capabilities. Within the framework of such a dimension, the management of the tools of the "mind" (self-criticism) becomes the most important knowledge for the agent, as well as competencies (self-reflection and skills (metacognition, as an accessible management tool). Despite the use of terminology that primarily implies living agents, such a structure can (and in many cases should) also characterize AI agents. The current focus on research on agency and self-awareness [58], in our opinion, is also closely related to anticipation, including when brain activity is modulated by internal signaling [59,60]. In this context, we can talk about intelligent systems in which there are other components that are different from the conventional "main" or "central brain", which, interacting with each other, give the system the very qualities of "hypercognition" that were discussed above.

It can also be argued that only the "power"/"computing power" of an intellectual system, does not, in itself, determine the "agency" and the connection with the temporal characteristics of the environment, the self-correlation of the "agent" with the processes occurring in time and anticipation. In other words, the ability to anticipate is probably an ability that does not directly depend on the conditional "strength" of the intellect, if, for example, one considers fish to be less intelligent in terms of "computational power" than a person. Conversely, one can consider that the strength of intellect can increase by summing simple organisms (a swarm of bees, a school of fish) or a set of artificial elements (for example, transistors. At the same time, the ability to predict is critical for living beings. With this approach, the ability to predict is what distinguishes a mere set of "intelligent" elements from a truly intelligent system.

## Event prediction and conditional environment constraints

Different factors affect the capacity of an agent to use available cognitive instruments. First, the objective uncertainty of the instruments. For example, in the case of evaluating predictive modeling [17] as an AI agent tool, one should take into account the structural, parametric, contextual uncertainty of the model, as well as the imperfection of input data and information (inaccuracy, inconsistency, incorrectness, integrity violations). As a result, the prediction obtained using such a model is associated with the uncertainty of the simulation results. Similarly, living agents (natural intelligence) rely on limited available tools, both external (available sources of information, means of communication, etc.) and internal (memory, ability to predict situations, physical skills, etc.). One can, however, objectively speak about the limitations of these tools and the uncertainty of their "output data".

Secondly, in people’s everyday activities, systems of "quick" reasoning are often used, which in their essence are heuristics based on simplification and usually work in most cases. This approach, being evolutionarily justified, gave rise, on the one hand, to a system of reflexes, and, on the other hand, led to such phenomena as cognitive distortions [61], irrational behavior [62] even in areas that at first glance require the priority of rational reasoning (economics, medicine, etc.). Nevertheless, at the level of meta-cognitive reasoning, the assessment of internal cognitive abilities is critical for reliable and informed decision-making. Finally, social interaction in teams of various sizes has a significant impact on the work of intelligent agents. Meanwhile, collective behavior in a sense is also a heuristic that simplifies the reasoning of an individual agent and expands the available tools to the level of distributed intelligence. On the one hand, a distributed intelligent system is able to store a larger amount of knowledge, perceive and process a larger amount of information and, as a result, solve more complex problems [63]. On the other hand, such large-scale systems have increased complexity [64,65] and are characterized by the appearance of emergent properties. As a result, the systemic effect can lead not only to positive [63], but also to negative effects [66]. . Moreover, a system with emergent properties may function optimally from the point of view of the system as a whole, but not optimally from the point of view of a single agent. And again, this can characterize the system positively or negatively, depending on the ratio of collective and individual good. The individual agent is under the influence of all three constraints: instrumental-objective, subjective, and collective (inter-subjective: from the point of view of the agent). At the same time, a decision-oriented intelligent agent must adequately control the operation of the available tools.On the one hand, AI expands the number of available tools, but at the current level of development, we can talk about distributed intelligent systems [67], in which natural and artificial intelligence agents operate simultaneously, each of which has its own set of (limited) tools, and (limited) cognitive abilities.

At the same time, AI becomes an element similar to a person, which means that it is required to have a similar level of responsibility, adequacy and even ethics. On the other hand, one must consider that, due to the emergent properties of the system on a large scale, “naturally ethical” people and “artificially ethical” AI can generate systemic effects that deviate from the established rules (including ethics). In order to prevent this from happening, it is necessary to take into account the hybrid, symbiotic nature of distributed intelligence systems that are emerging now and may appear in the future. This approach will allow to form the emergent properties of distributed intelligent systems in the "automatic mode" even in a situation of weak control at the level of the functional properties of individual agents.

With regard to a complex environment, such as, for example, human society, it will always be difficult for a real intellectual system (a living person or a hypothetical artificial intelligence) to choose an intelligent solution to fulfill the super-conditions specified by the current ethics. Similar paradoxes are described by Isaac Asimov in the fantastic construction of "laws of robotics", when there is a problem of choice, for example, between the protection of the identity of the robot, the human owner, and other people.

## Dual-axis intelligent system assessment scale

For a correct assessment of cognitive features, a non-one-dimensional scale can be proposed. One of the dimensions of this scale is the level of available cognitive tools (conditional "computing power", "performance"), and the other is the degree of adequate control of the tool by the forecasting agent. The advantage of such a scale is the possibility of a unified display of various systems: AI agents, live agents, distributed systems, etc. Considering such a scale (Fig. 2), you can display the following on it, in descending order of performance and increasing order of adequacy:

1) algorithmic solutions with predefined logic;
2) AI algorithms on data, including algorithms for machine learning, data mining, etc.;
3) "classical" AI algorithms based on symbolic knowledge;
4) mass intelligent systems (as an ideal abstraction of "superconsciousness" in the terminology of ConsScale [ConScale]).

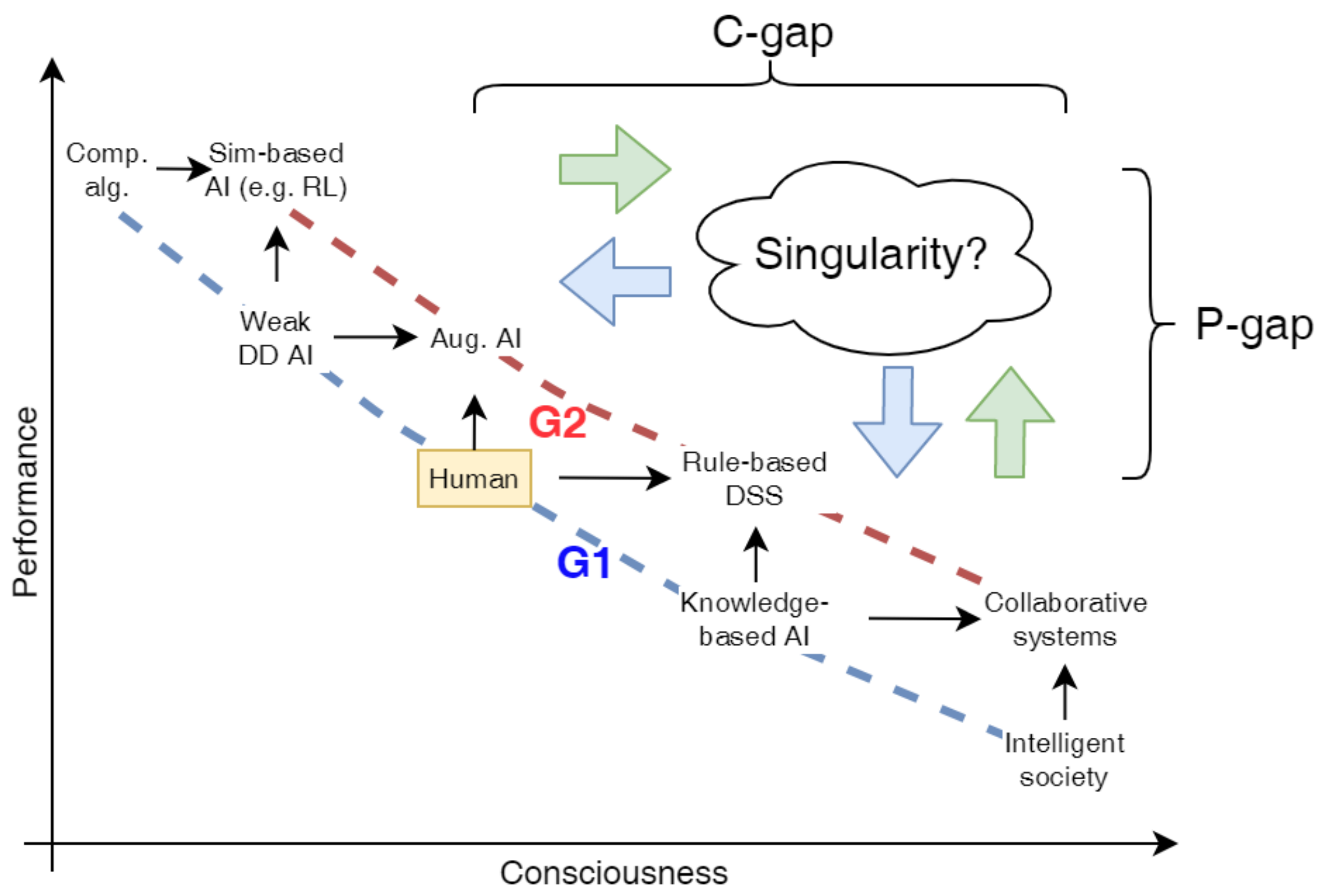

Figure 2 – Dual-axis intelligent system assessment scale

A person within this scale is in an "average" position, inferior in performance to computer systems (1-2), but significantly surpassing them in the mastery of available tools. Meanwhile, algorithms based on knowledge (3) are on the other side of the scale, because, due to a limited knowledge base, they have a smaller "outlook", but often greater stability and interpretability, which can be considered as a level of awareness. While mass intelligent systems (4) have common conditional knowledge/consciousness, they also exhibit significant inconsistency.

These systems can be considered relatively homogeneous. Moreover, it is possible to consider the hybridization of such systems, which in many situations makes it possible to build the second (G2) generation of intelligent systems, which can, in the long term, surpass the first (G1). Thus:

- If we combine deterministic algorithms with algorithms on data (1 +2), we get solutions such as reinforcement learning [68] and, in particular, imitation learning [69];
- if we combine AI algorithms based on data and human intelligence (2+H), an actual area of "augmented intelligence" (augmented AI) or human-centered AI (human-ceneterd AI) is formed [70,71];
- within the framework of combining human agents and knowledge-based systems (H + 3), scenarios of human-machine interaction arise, which are also within the framework of expert systems, decision support systems, etc.;
- combining knowledge-based systems with large-scale systems (3 + 4) corresponds to knowledge exchange systems, distributed learning systems, etc. [72,73].

It should be noted that the solutions of the second generation seem to be more promising and relevant today in terms of AI development and building information systems. Nevertheless, even within the framework of a multidimensional scale, it remains possible to improve both the performance of the tools available to agents (P-gap) and the appropriateness of their use (C-gap). In the future, controllable complex systems, having emergence, can create precedents for self-organizing distributed intelligent systems [67] with desired or unexpected properties.

This scale can be augmented with other dimensions. An obvious candidate for a third dimension can be the mass character (number) of intelligent systems. Large-scale mass systems can be present at various levels:

- computing systems [74] and workflow management systems) [75] can serve as an example for the level of deterministic algorithms;
- at the level of AI algorithms on data (2) this can be distributed data processing systems [76], including Internet of Things (IoT) systems [77];
- at the human level (H) this can be social systems;
- at the level of knowledge systems this can be systems of intelligent agents [78];
- at the level of mass systems this can be multi-agent systems, including hybrid, multi-agent reinforcement learning systems, etc, [79].

However, when subjecting intelligent systems to an increasingly sophisticated classification, two conditional axes should be considered basic: "brute" power (productivity) and "fine" power (adequacy, anticipation). These two axes mainly determine the integral and contextual effectiveness of an intelligent system. Thus, the two key properties that characterize the "mind" of an intellectual system are, on one hand, "power" and "productivity", and on the other, "adequacy" and"anticipativity"; brute and intelligent power at the same time.

## Conclusion

Constantly within the time stream, we often take the flow of time from the past to the future for granted, as well as the development of physical, chemical, biological and other processes in time. When evaluating computers,we only focus on computing speed: this many calculations per second. However, the key property of a true intelligent system is the ability to correlate the solution being prepared with the passage of time, with the same time flow within which the system is located. This can be likened to the calculation of future moves in chess. However, chess has very little predictive complexity, compared to the real challenges that even small fish face, taking into account conditions and options. For an intellectual system, this quality can also be conditionally designated as "range of vision" in time of changing conditions, the ability to anticipate. In the dual-axis scale we propose, this "range of vision" is represented as a modulating, "subtle" force that affects the achievable efficiency of "brute" power, which in turn is represented as "performance" or "computing power". At the same time, the "range of vision" can also be designated as "adequacy", correspondence to the context, temporal characteristics of what is known. The ability to anticipate is inherent even in the simplest living organisms, for example, protozoa capable of escaping drying out by crawling towards moisture, or crawling from light to shade, or vice versa. Taking into account that anticipation is not necessarily associated with a gigantic individual "computing power" (developed brain), we highlight it as a separate quality of an intellectual system. Now, if we evaluate an intelligent system not only by the number of transistors, calculations per second, and other attributes of "brute" force, but also by its relationship with the environment (conditions) that changes over time, we can get a more accurate, understandable characteristic of the "mind" of such a system. Thus, we are approaching the concept of common sense.

## References

[1] S.G. da Silva, J.J. Tehrani, Comparative phylogenetic analyses uncover the ancient roots of Indo-European folktales, R. Soc. Open Sci. 3 (2016) 150645. https://doi.org/10.1098/rsos.150645.

[2] C. Noone, M.J. Hogan, A randomised active-controlled trial to examine the effects of an online mindfulness intervention on executive control, critical thinking and key thinking dispositions in a university student sample, BMC Psychol. 6 (2018) 13. https://doi.org/10.1186/s40359-018-0226-3.

[3] Волшебная чаша. Индийские сказки в переводе Н. Ходзы, Детгиз, Л., 1956.

[4] П.К. Анохин, Биология и нейрофизиология условного рефлекса, Медицина, М., 1968.
[5] N. Wiener, Cybernetics: Or Control and Communication in the Animal and the Machine, 1948.
[6] M. Maron, Design principles for an intelligent machine, IEEE Trans. Inf. Theory. 8 (1962) 179–185. https://doi.org/10.1109/TIT.1962.1057751.
[7] T. Macpherson, A. Churchland, T. Sejnowski, J. DiCarlo, Y. Kamitani, H. Takahashi, T. Hikida, Natural and Artificial Intelligence: A brief introduction to the interplay between AI and neuroscience research, Neural Networks. 144 (2021) 603–613. https://doi.org/10.1016/j.neunet.2021.09.018.
[8] E. Dubé, D. Gagnon, E. Nickels, S. Jeram, M. Schuster, Mapping vaccine hesitancy—Country-specific characteristics of a global phenomenon, Vaccine. 32 (2014) 6649–6654. https://doi.org/10.1016/j.vaccine.2014.09.039.
[9] K.H. Tram, S. Saeed, C. Bradley, B. Fox, I. Eshun-Wilson, A. Mody, E. Geng, Deliberation, Dissent, and Distrust: Understanding Distinct Drivers of Coronavirus Disease 2019 Vaccine Hesitancy in the United States, Clin. Infect. Dis. 74 (2022) 1429–1441. https://doi.org/10.1093/cid/ciab633.
[10] B. Arul, D. Lee, S. Marzen, A Proposed Probabilistic Method for Distinguishing Between Delusions and Other Environmental Judgements, With Applications to Psychotherapy, Front. Psychol. 12 (2021) 674108. https://doi.org/10.3389/fpsyg.2021.674108.
[11] M. Foucault, Madness and civilization: A history of insanity in the age of reason, 2001.
[12] T.R. Shultz, The Bayesian revolution approaches psychological development, Dev. Sci. 10 (2007) 357–364. https://doi.org/10.1111/j.1467-7687.2007.00588.x.
[13] V.I. Arnold, Catastrophe Theory, Springer Berlin Heidelberg, Berlin, Heidelberg, 1992. https://doi.org/10.1007/978-3-642-58124-3.
[14] P. Bak, C. Tang, K. Wiesenfeld, Self-organized criticality: An explanation of the 1/ f noise, Phys. Rev. Lett. 59 (1987) 381–384. https://doi.org/10.1103/PhysRevLett.59.381.
[15] E. Ott, Chaos in Dynamical Systems, Cambridge University Press, 2002.
[16] B.R.R. Boaretto, R.C. Budzinski, K.L. Rossi, T.L. Prado, S.R. Lopes, C. Masoller, Discriminating chaotic and stochastic time series using permutation entropy and artificial neural networks, Sci. Rep. 11 (2021) 15789. https://doi.org/10.1038/s41598-021-95231-z.
[17] W. Walker, P. Harremoës, J. Rotmans, J.P. van der Sluijs, M.B.A. van Asselt, P. Janssen, M.P. Krayer von Krauss, Defining Uncertainty: A Conceptual Basis for Uncertainty Management in Model-Based Decision Support, Integr. Assess. 4 (2003) 5–17. https://doi.org/10.1076/iaij.4.1.5.16466.
[18] W.J. Murdoch, C. Singh, K. Kumbier, R. Abbasi-Asl, B. Yu, Definitions, methods, and applications in interpretable machine learning, Proc. Natl. Acad. Sci. 116 (2019) 1–11. https://doi.org/10.1073/pnas.1900654116.
[19] A. Abdul, J. Vermeulen, D. Wang, B.Y. Lim, M. Kankanhalli, Trends and Trajectories for Explainable, Accountable and Intelligible Systems, in: Proc. 2018 CHI Conf. Hum. Factors Comput. Syst., ACM, New York, NY, USA, 2018: pp. 1–18. https://doi.org/10.1145/3173574.3174156.
[20] G. Diana, T.T.J. Sainsbury, M.P. Meyer, Bayesian inference of neuronal assemblies, PLOS Comput. Biol. 15 (2019) e1007481. https://doi.org/10.1371/journal.pcbi.1007481.
[21] A. Kutschireiter, S.C. Surace, H. Sprekeler, J.-P. Pfister, Nonlinear Bayesian filtering and learning: a neuronal dynamics for perception, Sci. Rep. 7 (2017) 8722. https://doi.org/10.1038/s41598-017-06519-y.
[22] J. Pearl, D. Mackenzie, The book of why: the new science of cause and effect, 2018.
[23] Y. Antonacci, L. Minati, L. Faes, R. Pernice, G. Nollo, J. Toppi, A. Pietrabissa, L. Astolfi, Estimation of Granger causality through Artificial Neural Networks: applications to physiological systems and chaotic electronic oscillators., PeerJ. Comput. Sci. 7 (2021) e429. https://doi.org/10.7717/peerj-cs.429.
[24] A. Herzig, E. Lorini, J.F. Hubner, L. Vercouter, A logic of trust and reputation, Log. J. IGPL. 18 (2010) 214–244. https://doi.org/10.1093/jigpal/jzp077.
[25] C. Allen, I. Smit, W. Wallach, Artificial Morality: Top-down, Bottom-up, and Hybrid Approaches, Ethics Inf. Technol. 7 (2005) 149–155. https://doi.org/10.1007/s10676-006-0004-4.
[26] E. Santow, Emerging from AI utopia, Science (80-. ). 368 (2020) 9–9. https://doi.org/10.1126/science.abb9369.
[27] M. El Halabi, S. Mitrović, A. Norouzi-Fard, J. Tardos, J. Tarnawski, Fairness in Streaming Submodular Maximization: Algorithms and Hardness, (2020) arXiv:2010.07431.
[28] X. Wang, S. Wang, P.-Y. Chen, Y. Wang, B. Kulis, X. Lin, S. Chin, Protecting Neural Networks with Hierarchical Random Switching: Towards Better Robustness-Accuracy Trade-off for Stochastic Defenses, in: Proc. Twenty-Eighth Int. Jt. Conf. Artif. Intell., International Joint Conferences on Artificial Intelligence Organization, California, 2019: pp. 6013–6019. https://doi.org/10.24963/ijcai.2019/833.
[29] I. Gabriel, Artificial Intelligence, Values, and Alignment, Minds Mach. 30 (2020) 411–437. https://doi.org/10.1007/s11023-020-09539-2.
[30] S. V. Kovalchuk, G.D. Kopanitsa, I. V. Derevitskii, G.A. Matveev, D.A. Savitskaya, Three-stage intelligent support of clinical decision making for higher trust, validity, and explainability, J. Biomed. Inform. 127 (2022) 104013. https://doi.org/10.1016/j.jbi.2022.104013.
[31] D. Wang, J.D. Weisz, M. Muller, P. Ram, W. Geyer, C. Dugan, Y. Tausczik, H. Samulowitz, A. Gray, Human-AI Collaboration in Data Science, Proc. ACM Human-Computer Interact. 3 (2019) 1–24. https://doi.org/10.1145/3359313.
[32] W. Geyer, J. Weisz, C.S. Pinhanez, E. Daly, What is human-centered AI?, (2022). https://research.ibm.com/blog/what-is-human-centered-ai.
[33] N. Zheng, Z. Liu, P. Ren, Y. Ma, S. Chen, S. Yu, J. Xue, B. Chen, F. Wang, Hybrid-augmented intelligence: collaboration and cognition, Front. Inf. Technol. Electron. Eng. 18 (2017) 153–179. https://doi.org/10.1631/FITEE.1700053.
[34] D.K.I. Weidele, J.D. Weisz, E. Oduor, M. Muller, J. Andres, A. Gray, D. Wang, AutoAIViz: opening the blackbox of automated artificial intelligence with conditional parallel coordinates, in: Proc. 25th Int. Conf. Intell. User Interfaces, ACM, New York,

NY, USA, 2020: pp. 308–312. https://doi.org/10.1145/3377325.3377538.
[35] P. Cavalin, V.H.A. Ribeiro, M. Vasconcelos, C. Pinhanez, J. Nogima, H. Ferreira, Towards a Method to Classify Language Style for Enhancing Conversational Systems, in: 2021 Int. Jt. Conf. Neural Networks, IEEE, 2021: pp. 1–8. https://doi.org/10.1109/IJCNN52387.2021.9534090.
[36] D. Wang, J. Andres, J.D. Weisz, E. Oduor, C. Dugan, AutoDS: Towards Human-Centered Automation of Data Science, in: Proc. 2021 CHI Conf. Hum. Factors Comput. Syst., ACM, New York, NY, USA, 2021: pp. 1–12. https://doi.org/10.1145/3411764.3445526.
[37] D. Wang, Q. Yang, A. Abdul, B.Y. Lim, Designing Theory-Driven User-Centric Explainable AI, in: Proc. 2019 CHI Conf. Hum. Factors Comput. Syst., ACM, New York, NY, USA, 2019: pp. 1–15. https://doi.org/10.1145/3290605.3300831.
[38] Ж. Делёз, Ф. Гваттари, Что такое философия?, Алетейя, СПб., 1998.
[39] А.М. Буровский, Очередная планетарная революция или уникальная сингулярность?, Историческая Психология и Социология Истории. (2018) 62.
[40] А.П. Назаретян, Л.А. Карнацкая, Историко-психологическая подоплека глобальных вызовов, Развитие Личности. (2017) 20–46.
[41] А.Н. Аверкин, С.А. Ярушев, В.Ю. Павлов, Когнитивные гибридные системы поддержки принятия решений и прогнозирования, Программные Продукты и Системы. 30 (2017) 632–642. https://doi.org/10.15827/0236-235X.030.4.632-642.
[42] C.-C. Chen, C.-C. Wei, S.-H. Chen, L.-M. Sun, H.-H. Lin, AI Predicted Competency Model to Maximize Job Performance, Cybern. Syst. 53 (2022) 298–317. https://doi.org/10.1080/01969722.2021.1983701.
[43] D. Vrontis, M. Christofi, V. Pereira, S. Tarba, A. Makrides, E. Trichina, Artificial intelligence, robotics, advanced technologies and human resource management: a systematic review, Int. J. Hum. Resour. Manag. 33 (2022) 1237–1266. https://doi.org/10.1080/09585192.2020.1871398.
[44] K. Grünenberg, P. Møhl, K.F. Olwig, A. Simonsen, Issue Introduction: IDentities and Identity: Biometric Technologies, Borders and Migration, Ethnos. 87 (2022) 211–222. https://doi.org/10.1080/00141844.2020.1743336.
[45] M. Каку, Давайте станем суперлюдьми, (n.d.). https://hbr-russia.ru/innovatsii/tekhnologii/771054/.
[46] S.J. Russell, Human Compatible: Artificial Intelligence and the Problem of Control, 2019.
[47] A. Abbott, Inside the mind of an animal, Nature. 584 (2020) 182–185. https://doi.org/10.1038/d41586-020-02337-x.
[48] Л.В. Крушинский, Биологические основы рассудочной деятельности. Эволюционный и физиолого-генетический аспекты поведения, Изд-во Моск. ун-та, Москва, 1977.
[49] M. Nadin, Anticipation and Medicine, Springer International Publishing, Cham, 2017. https://doi.org/10.1007/978-3-319-45142-8.
[50] P.K. Anokhin, Nodular Mechanism of Functional Systems as a Self-regulating Apparatus, Prog. Brain Res. 22 (1968) 230–251. https://doi.org/10.1016/S0079-6123(08)63509-8.
[51] N.A. Bernstein, The Co-Ordination and Regulation of Movements, Pergamon Press, Oxford, 1967.
[52] M.G. Salena, A.J. Turko, A. Singh, A. Pathak, E. Hughes, C. Brown, S. Balshine, Understanding fish cognition: a review and appraisal of current practices, Anim. Cogn. 24 (2021) 395–406. https://doi.org/10.1007/s10071-021-01488-2.
[53] BICA*AI, Hierarchy of Levels of Cognition, (n.d.). https://bica.ai/hierarchy-of-levels-of-cognition/.
[54] R. Arrabales, Conscious-Robots.com | ConsScale - A Machine Consciousness Scale, (n.d.). https://www.conscious-robots.com/consscale/levels.
[55] I. Kotseruba, J.K. Tsotsos, 40 years of cognitive architectures: core cognitive abilities and practical applications, Artif. Intell. Rev. (2018). https://doi.org/10.1007/s10462-018-9646-y.
[56] S. Hélie, R. Sun, Incubation, insight, and creative problem solving: A unified theory and a connectionist model., Psychol. Rev. 117 (2010) 994–1024. https://doi.org/10.1037/a0019532.
[57] A. Fedor, I. Zachar, A. Szilágyi, M. Öllinger, H.P. de Vladar, E. Szathmáry, Cognitive Architecture with Evolutionary Dynamics Solves Insight Problem, Front. Psychol. 8 (2017) 1–15. https://doi.org/10.3389/fpsyg.2017.00427.
[58] P. Haggard, Sense of agency in the human brain, Nat. Rev. Neurosci. 18 (2017) 196–207. https://doi.org/10.1038/nrn.2017.14.
[59] O. Kubryak, The Anticipating Heart, in: Anticip. Med., Springer International Publishing, Cham, 2017: pp. 49–65. https://doi.org/10.1007/978-3-319-45142-8_4.
[60] A. Koreki, D. Goeta, L. Ricciardi, T. Eilon, J. Chen, H.D. Critchley, S.N. Garfinkel, M. Edwards, M. Yogarajah, The relationship between interoception and agency and its modulation by heartbeats: an exploratory study, Sci. Rep. 12 (2022) 13624. https://doi.org/10.1038/s41598-022-16569-6.
[61] E. O'Sullivan, S. Schofield, Cognitive Bias in Clinical Medicine, J. R. Coll. Physicians Edinb. 48 (2018) 225–232. https://doi.org/10.4997/jrcpe.2018.306.
[62] G.S. Becker, Irrational Behavior and Economic Theory, J. Polit. Econ. 70 (1962) 1–13. https://doi.org/10.1086/258584.
[63] J. Surowiecki, The Wisdom of Crowds: Why the Many Are Smarter Than the Few and How Collective Wisdom Shapes Business, Economies, Societies and Nations, 2004.
[64] H. Sayama, Introduction to the Modeling and Analysis of Complex Systems, Open SUNY Textbooks, New York, 2015.
[65] N. Boccara, Modeling Complex Systems, Springer New York, New York, 2010. https://doi.org/10.1007/978-1-4419-6562-2.
[66] C. Mackay, Extraordinary Popular Delusions and the Madness of Crowds, 1841.
[67] V. Guleva, E. Shikov, K. Bochenina, S. Kovalchuk, A. Alodjants, A. Boukhanovsky, Emerging Complexity in Distributed Intelligent Systems, Entropy. 22 (2020) 1437. https://doi.org/10.3390/e22121437.
[68] R.S.S. Sutton, A.G.G. Barto, Reinforcement Learning: An Introduction, 2nd ed., IEEE, 2012.
[69] A. Hussein, M.M. Gaber, E. Elyan, C. Jayne, Imitation Learning: A Survey of Learning Methods, ACM Comput. Surv. 50 (2018) 1–35. https://doi.org/10.1145/3054912.
[70] D. Glowacka, A. Howes, J.P. Jokinen, A. Oulasvirta, Ö. Şimşek, RL4HCI: Reinforcement Learning for Humans, Computers,

and Interaction, in: Ext. Abstr. 2021 CHI Conf. Hum. Factors Comput. Syst., ACM, New York, NY, USA, 2021: pp. 1–3. https://doi.org/10.1145/3411763.3441323.
[71] D. Harris, V. Duffy, M. Smith, C. Stephanidis, eds., Human-Centered Computing, CRC Press, 2019. https://doi.org/10.1201/9780367813369.
[72] L. Zhen, Z. Jiang, H. Song, Distributed recommender for peer-to-peer knowledge sharing, Inf. Sci. (Ny). 180 (2010) 3546–3561. https://doi.org/10.1016/j.ins.2010.05.036.
[73] M. Alavi, G.M. Marakas, Y. Yoo, A Comparative Study of Distributed Learning Environments on Learning Outcomes, Inf. Syst. Res. 13 (2002) 404–415. https://doi.org/10.1287/isre.13.4.404.72.
[74] R. Buyya, S.M. Thampi, eds., Intelligent Distributed Computing, Springer International Publishing, Cham, 2015. https://doi.org/10.1007/978-3-319-11227-5.
[75] J. Liu, E. Pacitti, P. Valduriez, M. Mattoso, A Survey of Data-Intensive Scientific Workflow Management, J. Grid Comput. 13 (2015) 457–493. https://doi.org/10.1007/s10723-015-9329-8.
[76] H. Isah, T. Abughofa, S. Mahfuz, D. Ajerla, F. Zulkernine, S. Khan, A Survey of Distributed Data Stream Processing Frameworks, IEEE Access. 7 (2019) 154300–154316. https://doi.org/10.1109/ACCESS.2019.2946884.
[77] S. Li, L. Da Xu, S. Zhao, The internet of things: a survey, Inf. Syst. Front. 17 (2015) 243–259. https://doi.org/10.1007/s10796-014-9492-7.
[78] M. Wooldridge, N.R. Jennings, Intelligent agents: theory and practice, Knowl. Eng. Rev. 10 (1995) 115–152. https://doi.org/10.1017/S0269888900008122.
[79] K. Zhang, Z. Yang, T. Ba\csar, Multi-agent reinforcement learning: A selective overview of theories and algorithms, ArXiv Prepr. ArXiv1911.10635. (2019).